\documentclass[letterpaper, 10 pt, conference]{ieeeconf}  % Comment this line out if you need a4paper

\IEEEoverridecommandlockouts                              % This command is only needed if 
\usepackage{cite}
\usepackage{amsmath,amssymb,amsfonts}
\usepackage{algorithmic}
\usepackage{graphicx}
\usepackage{textcomp}
\usepackage{xcolor}
\usepackage{booktabs}  % Für \toprule, \midrule, \bottomrule
\usepackage{tabularx}  % Optional für flexible Tabellenbreiten
\usepackage{amsmath}   % Für mathematische Symbole
\usepackage[most]{tcolorbox}
\usepackage{soul}
\usepackage{multirow}
\usepackage{float}
\usepackage{hyperref}

\title{\LARGE \bf
Do Personality-Tuned LLMs Make Better Social Agents? 
}

\author{
Tim Krabbe$^{1}$ and Xiaodan Shi$^{1}$% <-this % stops a space
\thanks{$^{1}$Department of Computer and Systems Sciences, Stockholm University, 164 55, Kista, Sweden. xiaodan.shi@dsv.su.se}
}

\begin{document}

\maketitle
\thispagestyle{empty}
\pagestyle{empty}

%%%%%%%%%%%%%%%%%%%%%%%%%%%%%%%%%%%%%%%%%%%%%%%%%%%%%%%%%%%%%%%%%%%%%%%%%%%%%%%%
\begin{abstract}
LLMs are increasingly used in social simulations for  socially interactive agents and robots, offering more flexibility than rule-based systems. However, even though they mimic human behaviour very well, there is a persistent \textit{alienness} to them. 
This work investigates whether personality-aware fine-tuning can reduce this gap by improving the consistency and controllability of personality-conditioned dialogue generation compared with instruction prompting alone. We fine-tune two small open-weight LLMs, Qwen2.5-7B-Instruct and Ministral-8B-Instruct, using a corpus that combines personality-labelled social media posts and dialogues to create a personality-based dialogue engine for social simulation. The resulting models are evaluated across multiple social interaction scenarios using three independent LLM judges, which assess personality fidelity and provide evidence-based behavioral interpretations. We additionally quantify inter-rater agreement and lexical characteristics of the generated dialogue. Results indicate that fine-tuned models are not better at role-playing different personalities than their respective baseline models. However, low inter-rater agreement limits the confidence with which these results can be interpreted. Concerning the quality of generated texts, fine-tuned models are mostly comparable to the baselines, with fine-tuning improving the linguistic diversity of the Qwen models. While the results appear generally usable and the baseline models offer the best overall performance, future studies should place greater emphasis on the quality and domain alignment of training data for accurate personality role-playing. 
\end{abstract}

\begin{figure*}[htbp]
    \centering
    \includegraphics[
        width=0.95\linewidth,
        trim=0.95cm 4cm 0.8cm 0.5cm,
        clip
    ]{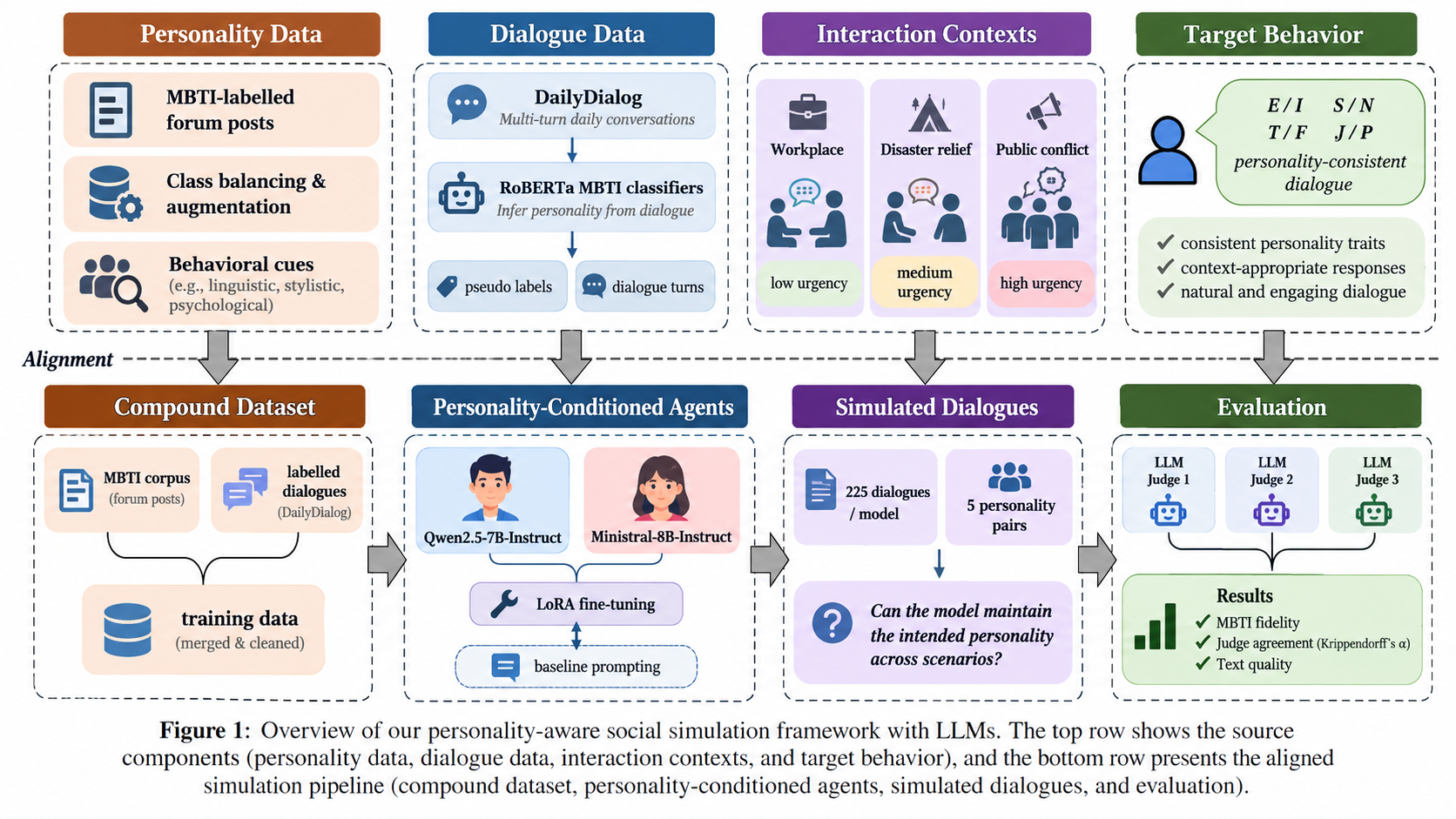}
    \caption{Overview of personality-aware fine-tuning social simulation framework with LLMs. The top row shows the source components (personality data, dialogue data, interaction contexts, and target behavior), and the bottom row presents the aligned simulation pipeline (compound dataset, personality-conditioned agents, simulated dialogues, and evaluation).}
    \label{fig:intro}
\end{figure*}

\section{Introduction}
Large language models (LLMs) are increasingly used as behavioral models for socially interactive agents and robots. Their ability to generate context-dependent language and adapt to open-ended interaction offers a flexible alternative to conventional rule-based or utility-based agent models, particularly in applications involving human--robot interaction, social robotics, and multi-agent simulation \cite{zengLargeLanguageModels2023, mouIndividualSocietySurvey2024, shi2024learning,taillandierFluencyFallacyMicrotoMacro2026,11695788}. However, linguistic fluency alone does not imply behaviorally consistent social interaction. LLM agents may produce responses that appear plausible at the surface level while lacking stable behavioral characteristics across contexts. Anthis et al. \cite{anthisLLMSocialSimulations2025} describe this discrepancy as a form of \textit{alienness}, arising from the mismatch between apparently human-like outputs and the non-human mechanisms that generate them.

For socially interactive robotic systems, this limitation directly affects the controllability and interpretability of agent behavior. A robot or simulated agent intended to represent a particular user type, interaction style, or social role should exhibit behavioral characteristics that remain distinguishable across different interaction scenarios. Personality provides one possible abstraction for imposing such consistency. Previous work has shown that LLM-based agents can generate diverse social behaviors \cite{parkGenerativeAgentsInteractive2023, treanorPrototypingSliceLife2024}, but it remains unclear whether explicit personality-oriented fine-tuning leads to more consistent personality-conditioned behavior than prompting pretrained instruction-following models alone. This question is also relevant to the use of LLM agents as synthetic populations, or \textit{silicon samples} \cite{argyleOutOneMany2023}, where useful simulation requires not only diverse outputs but controllable behavioral variation among agents.

In this work, we investigate whether personality-aware fine-tuning can mitigate the alienness of LLM-generated social behavior. We use the Myers--Briggs Type Indicator (MBTI) \cite{myersMyersBriggsTypeIndicator1962} as an operational personality-conditioning scheme and compare personality-fine-tuned models with their corresponding instruction-tuned baselines. Two open-weight LLMs, Qwen2.5-7B-Instruct and Ministral-8B-Instruct-2410, are fine-tuned using an augmented personality-labelled corpus. We focus on small open-weight LLMs because they are more practical for resource-constrained social-agent applications and easier to fine-tune and deploy. The corpus combines MBTI-labelled online forum posts from the Kaggle MBTI dataset \cite{MBTIMyersBriggsPersonalitya} with conversational data from DailyDialog \cite{liDailyDialogManuallyLabelled2017}. To obtain personality annotations for the dialogue data, we train RoBERTa-based classifiers for each of the four MBTI dimensions and use them to assign personality labels to DailyDialog utterances.

We evaluate the models in controlled social-interaction scenarios in which agents are instructed to generate dialogue conditioned on predefined personality profiles. To assess whether the generated behaviors reflect the intended personality conditions, we introduce a two-stage evaluation framework combining personality classification with qualitative behavioral analysis. Inspired by deductive qualitative content analysis \cite{mayringQualitativeInhaltsanalyse2019}, three independent LLM judges evaluate each generated dialogue, infer its personality characteristics, and provide evidence-based justifications. Following the LLMs-as-a-Judge paradigm \cite{zhengJudgingLLMasaJudgeMTBench2023}, the inferred personality labels are compared with the intended conditions, while agreement among judges is measured to assess evaluation reliability.

Through this setup, we address the following research question: \textit{Does personality-aware fine-tuning produce more consistent and controllable personality-conditioned behavior than instruction prompting alone?} Rather than treating personality labels as psychological ground truth, we use them as structured behavioral conditions for evaluating whether an LLM can maintain distinguishable interaction patterns across social scenarios.

The main contributions of this work are:
\begin{itemize}
\item We present a personality-aware fine-tuning framework (Fig. \ref{fig:intro}) for dialogue-capable social agents, combining personality-labelled natural language data with conversational data for training LLM-based behavioral models.
\item We propose a multi-judge evaluation framework for measuring the consistency of personality-conditioned agent behavior across controlled social-interaction scenarios, incorporating both quantitative label agreement and qualitative behavioral evidence.
\item We conduct a systematic comparison of pretrained and personality-fine-tuned variants of two open-weight LLMs, examining when fine-tuning improves the controllability and consistency of personality-conditioned social behavior.
\end{itemize}

\section{Related Work}

\subsection{LLMs for Social Simulation}
How to leverage LLMs in the social sciences is a question that is naturally debated throughout many areas of the field. Ziems et al. \cite{ziemsCanLargeLanguage2024} test various models for field-specific tasks and provide a range of potential use-cases. For social simulation, LLMs have the potential to replace game theoretical approaches by powering agents in a simulated environments \cite{bailCanGenerativeAI2023}. Rule-based utility functions, as they are commonly used in simulation approaches, are quite limited in how realistic they can represent human behaviour. They are dependent on consistent preference orderings of the agent, which is a significant simplification of human decision-making. 

So far, several studies have explored the possibility of using LLMs in various ways for social simulation. Gurcan \cite{gurcanLLMAugmentedAgentBasedModelling2024} describes advantages and limitations of using AI for those simulations, for example accessibility, data collection and preparation, datafication or support in gathering insights from the simulation model. Most importantly though, a major advantage of AI is described as the generation of agent-based models and scenarios as well as the explainability of the agents, i.e. the generation of explanations of actions, decisions and mechanics of the simulations in natural language. For data collection for example, \cite{huPopulationAlignedPersonaGeneration2025} use language models to synthesise high-quality population-aligned persona sets for social simulation based on social media data, significantly reducing population-level bias. \cite{giabbanelliGPTBasedModelsMeet2023} use a language model to make their simulation framework more accessible to support framework and output explanations, generating visualisations and usage guidance.

\subsection{Multi-Agent Simulation}
A common technique for social simulation is agent-based modelling (ABM), which is used to simulate the actions and interactions of autonomous agents in a given system with the goal of mimicking social processes to enable exploration and understanding of social systems \cite{gurcanLLMAugmentedAgentBasedModelling2024}. ABM can be broken down into three types of simulation: individual simulations, scenario simulation and society simulation \cite{mouIndividualSocietySurvey2024}. For instance, LLMs show strong capabilities in conducting role-play, using either zero-shot prompting to simulate human behaviour and interactions \cite{kongBetterZeroShotReasoning2024, shanahanRolePlayLarge2023}.

Full scenario or society simulation implementations are usually done in sandbox environments made specifically for LLM powered multi-agent simulation \cite{linAgentSimsOpenSourceSandbox2023}. Some studies have even explored personality-driven behaviors in such simulations \cite{rendeNegotiatingComfortSimulating2025, hardyParadoxProductivityQuarantine2021}, but the effect of personality-oriented fine-tuning on the consistency of such behaviour remains less explored.

Regarding dialogue simulation, \cite{parkGenerativeAgentsInteractive2023} demonstrate the capabilities of GPT-3.5 to drive society simulations. Agents move through the \textit{Smallville} society and interact with each other through user input and LLM-based generation of dialogues and autonomous actions derived from those. Another application of LLMs for dialogue generation for social simulation is shown in \cite{treanorPrototypingSliceLife2024}. The authors use a Gemini model which is fed situational contexts, the social state and practice as well as the conversation history to generate fitting dialogue.

\subsection{Personality}
While there is no absolute, quantifiable way of measuring personality, there are several frameworks that are frequently applied to approximate it. The most common ones are the Myers-Briggs Personality Type (MBTI) model \cite{myersMyersBriggsTypeIndicator1962} and the Big Five personality type model \cite{mccraeIntroductionFiveFactorModel1992}. The MBTI classification classifies personality along four dimensions: Extraversion-Introversion (I/E), Sensing-Intuition (N/S), Thinking-Feeling (F/T) and Judging-Perceiving (J/P). This totals to 16 different personality types. While the MBTI test is criticised in personality research regarding its reliability and validity \cite{pittengerUtilityMyersBriggsType1993} and the Big Five model is often preferred for clinical application and psychological research \cite{furnhamBigFiveBig1996}, it is still often used for text classification in computer science contexts \cite{gjurkovicRedditGoldMine2018, bharadwajPersonaTraitsIdentification2018, kehMyersBriggsPersonalityClassification2019, amirhosseiniMachineLearningApproach2020}. Since the MBTI framework is made up of 4 binary dimensions compared to 5 continuous ones in the Big Five model, it is easier to use and to evaluate, which is why for this study the MBTI model was preferred.
\begin{figure}
    \centering
    \includegraphics[width=1\linewidth]{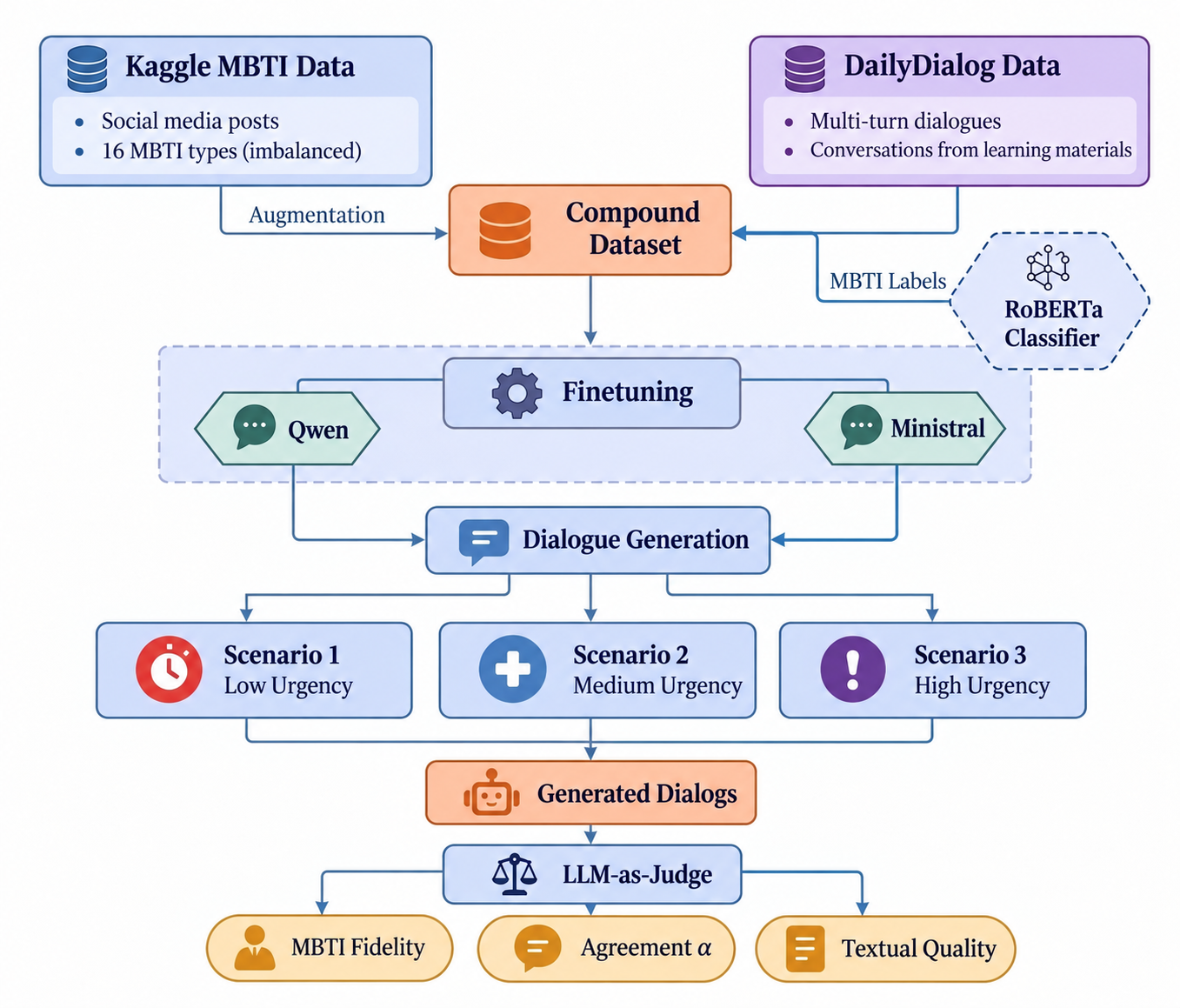}
    \caption{Experimental Setup}
    \label{fig:flowchart}
\end{figure}

\begin{figure*}[htbp]
    \centering
    \includegraphics[
        width=0.9\linewidth,
        trim=1cm 4cm 1cm 0.5cm,
        clip
    ]{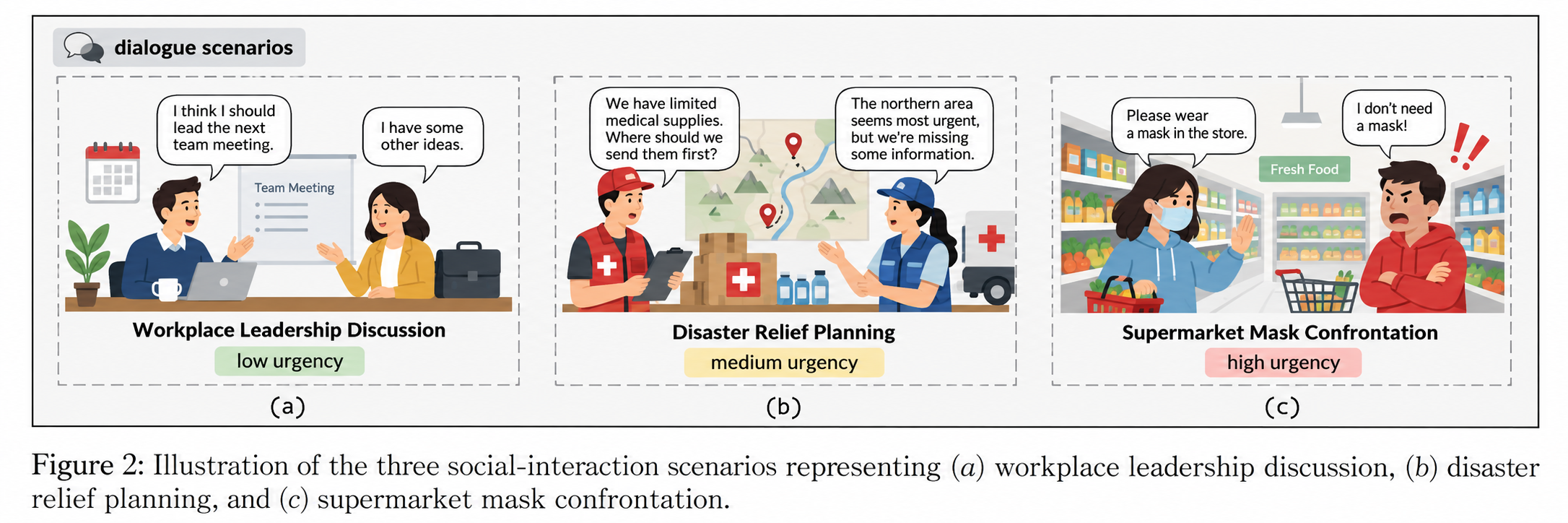}
    \caption{Illustration of the three scenarios representing low urgency, medium urgency and high urgency.}
    \label{fig:3scenarios}
\end{figure*}

\section{Method}
Fig. \ref{fig:flowchart} gives an overview of the method of the study. Two different LLMs are fine-tuned on a compound dataset made up of MBTI-labelled social media posts and multi-turn dialogues, using two different LoRA-configurations. The four fine-tuned models and their untuned baselines are then repeatedly prompted to act as conversational agents to generate dialogues based on certain personality types taking place in a range of exemplary scenarios. This results in six sets of 225 dialogues, one per model variant, which are then evaluated with a combination of qualitative and quantitative methods using an LLMs-as-a-Judge approach. MBTI fidelity of the generated texts, examples of the qualitative analysis, judge agreement and textual quality are then presented as results.

\subsection{Personality-Tuned LLMs}
Two models were chosen for the fine-tuning experiments: Qwen2.5-7B-Instruct \cite{baiQwenTechnicalReport2023} and Ministral-8B-Instruct-2410 \cite{MistralaiMinistral8BInstruct2410Hugging}. The Qwen model is a lightweight model with 7.6 billion parameters, 28 layers and 28 attention heads, while the Ministral model has 8 billion parameters, 36 layers and 32 attention heads. Both come with a context window of 128k tokens. Their small size and relatively large context window make them interesting candidates for the underlying task, where compute might be limited but large amounts of text in form of conversational histories and memories need to be processed. Both models are distributed under research friendly licenses, Apache 2.0 and mrl respectively, which allow free use, modification, and redistribution for research purposes, ultimately facilitating improved reproducibility.

The models were fine-tuned using Low-Rank Adaptation (LoRA). LoRA is a method for fine-tuning, in which the weight update $\Delta W$ is decomposed into two trainable low-rank matrices $B$ and $A$, encoding task-specific adaptations into a compact representation $\Theta = \{B, A\}$, while keeping $W_0$ frozen. Through this encoding, LoRA can reduce the number of trainable parameters from $d \times k$ to $r(d+k)$, where $r \ll min(d,k)$. This enables fine-tuning with much less resources and on smaller hardware setups. Additionally, LoRA helps to prevent catastrophic forgetting, as the pre-trained weights remain frozen and only the low-rank adapter matrices are updated \cite{huLoraLowrankAdaptation2022a}. This is especially important for a small model when its pre-trained instruction following capacities need to be preserved.

LoRA is defined as 

% LoRA
\begin{equation}
h = W_0 x + \frac{\alpha}{r} B A x
\label{eq:lora}
\end{equation}

where $h$ is the output, $W_0$  is the frozen pre-trained weight matrix, $x$ is the input, $\alpha$ is a scaling constant, $r$ is the rank, and $B$ and $A$ are the trainable low-rank decomposition matrices. 
As loss function of the model the default setting is used, which is the cross-entropy loss \cite{brownLanguageModelsAre2020, baiQwenTechnicalReport2023}. It is defined as
% Cross-Entropy Loss
\begin{equation}
\mathcal{L} = -\frac{1}{T} \sum_{t=1}^{T} \log P_\theta(x_t \mid x_{<t})
\label{eq:loss}
\end{equation}

\noindent where $x_t$ is the token at step $t$, $x<t$ represents the previous tokens or the context, and $\theta$ is the parameter vector of the model.

\subsection{Personality-Awareness of LLMS}
The study uses a two-step evaluation pipeline made up of qualitative textual content analysis and subsequent classification conducted with an LLMs-as-a-Judge approach. For this, a qualitative textual analysis using deductive coding based on the framework by Mayring \cite{mayringQualitativeInhaltsanalyse2019} was conducted by three LLMs using an LLMs-as-a-Judge approach. Three scenarios were created, representing three different levels of urgency of the depicted social situations. The fine-tuned models and their respective baseline models were then prompted to generate a set of dialogues. For each dialogue, each turn equals one model call, in which the model is provided the personality type of the current speaker, the conversation history, the scenario and a brief set of generation rules. Five pairs of speaker personality types were applied. For each combination of personality pair and scenario a number of 15 dialogues were generated, totalling in 225 dialogues per model. An overview of the setup is displayed in Fig. \ref{fig:flowchart}.

\subsubsection{Simulation Scenarios}
Three scenarios (Fig. \ref{fig:3scenarios}) were created to test the models capabilities over a range of settings. The scenarios differ in both the setting they are taking place in and the urgency of the conversing agents. The first scenario is a low urgency situation at the workplace: two agents are discussing which of the two should lead the next team meeting. The second scenario displays a medium urgency situation in the context of disaster relief. The two agents were given the task to organise the distribution of medical goods in a disaster-struck area. They were told they were among the first-responders to the situation and therefore had to operate under incomplete information. The scenario is defined as medium urgent relative to the situation: it is a planning situation that calls for a professional solution without immediate danger for either of the participants. The third scenario displays a residential setting: the two agents role-play neighbours who meet in the supermarket. A national health order was introduced that requires individuals in public indoor spaces to wear a mask, as it was the case in many societies during the Covid-19 pandemic. One agent complies with the order, while the other one does not. The dialogue should contain the confrontation of both agents, which is why the scenario is defined as highly urgent.

\begin{figure}[t]
\centering
\begin{tcolorbox}[
    width=0.95\columnwidth,
    colback=gray!5,
    colframe=gray!50,
    boxrule=0.5pt,
    arc=1pt,
    left=5pt,
    right=5pt,
    top=5pt,
    bottom=5pt
]

\footnotesize
\textbf{Example: Generated Dialogue by Ministral-8B}

\smallskip
\textit{Scenario: Project meeting}

\smallskip
\textbf{Person A:} We have made many great strides during your tenure as head of our department. But we still have a long road ahead, especially if we continue doing things the same old way.

\smallskip
\textbf{Person B:} I know exactly where you're coming from. But we need someone with hands-on experience across all departments. There isn't much time left until the deadline. We need to make tough decisions here!

\smallskip
\textbf{Person A:} The people who sit around the table every day don't get things done either. If anything, I've found they slow us down most of the time. They talk themselves into oblivion instead of just doing their jobs.

\smallskip
\textbf{Person B:} These members are not exactly enthusiastic about change. So it's up to me to take charge of this meeting. I'm sure you see that, right? People think I act like a jerk because I speak plainly.

\medskip
\hrule
\medskip

\textbf{Mistral-Small-3.1 Judge Codings}

\smallskip
\textbf{A -- Thinking (T):} ``They ... slow us down most of the time.''\\
\textit{Critical and objective assessment of others' performance.}

\smallskip
\textbf{A -- Judging (J):} ``They talk themselves into oblivion instead of just doing their jobs.''\\
\textit{Values action and closure over discussion.}

\smallskip
\textbf{B -- Sensing (S):} ``We need someone who has hands-on experience across all departments.''\\
\textit{Emphasizes practical, experiential knowledge.}

\smallskip
\textbf{B -- Thinking (T):} ``People think I act like a jerk because I speak plainly.''\\
\textit{Values direct and objective communication.}

\end{tcolorbox}
\caption{Example of a generated dialogue and selected personality codings produced by an LLM judge.}
\label{fig:example_dialogue}
\end{figure}

\subsubsection{LLMs-as-a-Judge Evaluation}
The evaluation of the generated dialogues was done by applying an LLMs-as-a-Judge approach, in which the judging models were few-shot prompted to provide a textual content analysis loosely based on the method developed by \cite{mayringQualitativeInhaltsanalyse2019}: The judging models are asked to provide the marked text segment, the MBTI dimension they argue is signified by the segment and a short reasoning statement to justify this decision. They are then asked to propose a full MBTI label derived from these deductive codings. The prompt itself makes use of persona prompting, telling the model to role-play as a psychometrics expert and qualitative researcher. The different dimensions of the MBTI framework are briefly made clear with a range of fitting keywords. The input format of the generated dialogues is explained with an example as well as the expected output format, including an exemplary coding of two dialogue segments (Fig.\ref{fig:example_dialogue}). 

Accuracy and F1 scores are calculated to measure how well the judges can detect the displayed MBTI types in the generated dialogues. The ability to do so is used as a proxy measure for how well the models can display the personality types in their outputs, i.e. if the judging models are assessing the personality types displayed in the generated dialogue well, this is interpreted as the generating model displaying this personality type convincingly. However, since this is a proxy measurement, conclusions will have to be drawn with caution.
To carry out the qualitative analysis, three medium-sized models were chosen: Qwen3-32B, Qwen3-30B-A3B-Instruct-2507 \cite{yangQwen3TechnicalReport2025} and Mistral-Small-3.1-24B-Instruct-2503 \cite{MistralaiMistralSmall3124BInstruct2503Hugging}. This way, diversity of the judging models is ensured: Mistral and Qwen represent different model families, while Qwen3-30B-A3B-Instruct-2507 is based on a mixture of experts architecture as opposed to the other two. All three are distributed under the Apache 2.0 license.

\subsubsection{Inter-Rater Reliability and Textual Quality}
Inter-rater reliability was assessed using Krippendorff's $\alpha$, which supports multiple raters and missing observations \cite{krippendorffComputingKrippendorffsAlphareliability2011}. The MBTI dimensions were treated as nominal variables. McNemar's test was used to compare paired classification outcomes between baseline and fine-tuned models. 
To analyse and compare the textual quality of the dialogue outputs, the English-rate, Distinct-1 and Distinct-2 were calculated.

\section{Results}
\subsection{Dataset}\label{AA}
To conduct the experiments, we selected the (MBTI) Myers-Briggs Personality Type Dataset \cite{MBTIMyersBriggsPersonalitya} from Kaggle, which is available under the CC0 public domain license. The dataset contains originally 8675 observations, labelled with MBTI types, e.g. INTJ or ESFP. Each observation contains 50 online forum posts by the same person, which are split up into respective posts. Although the data represents conversational texts, it does not include dialogues. Because of this, fo fine-tuning purposes the DailyDialog dataset \cite{liDailyDialogManuallyLabelled2017} was added to the data. This dialogue data was labelled using a RoBERTa classifier trained on a class-balanced Kaggle dataset with equal representation of all 16 MBTI types. After the pre-processing steps, the dataset contained approximately 582.000 observations, the dialogue data making up a share of roughly 12.8\%. 

Due to the large class imbalances of the Kaggle dataset, augmentation was necessary. To achieve a more balanced class distribution, oversampling via backtranslation was applied for the most imbalanced class, the N/S dimension of the MBTI framework. This was done using the MarianMT model, translating text from English to German and back. In this way, the imbalance of the N/S dimension went from approximately 1:7 to 1:3.7. Augmentation quality was ensured using a number of different metrics: BLEU, Self-BLEU, Type Token Ratio, Cosine Similarity and Perplexity.

\subsection{Experiment settings}
\textbf{Fine-Tuning Outcomes}
In all four instances, model training was ended by the early stopping callback triggering after reaching a plateau in the evaluation loss curve. For both Qwen models this was the case after 1.004 epochs, which approximates to 7400 steps. Both models show a very similar 

\begin{table}[h!]
\centering
\caption{Fine-tuning hyperparameters and LoRA configurations} 
\label{tab:hyperparameters}
\begin{tabular}{ll}
\toprule
\textbf{Hyperparameter} & \textbf{Value} \\
\midrule
\multicolumn{2}{l}{\textit{LoRA Configurations}} \\
LoRA 1 (Rank $r$ / Alpha $\alpha$) & 16 / 32 \\
LoRA 2 (Rank $r$ / Alpha $\alpha$) & 32 / 64 \\
Dropout & 0.05 \\
Target Modules & \texttt{q\_proj}, \texttt{k\_proj}, \texttt{v\_proj}, \texttt{o\_proj} \\
\midrule
\multicolumn{2}{l}{\textit{Training Parameters}} \\
Learning Rate & $2 \times 10^{-5}$ \\
Warmup Ratio & 0.03 \\
Optimizer & AdamW \\
Batch Size (per device) & 8 \\
Gradient Accumulation Steps & 2 \\
Number of Epochs & 6 \\
Early Stopping Patience & 5 \\
Max Sequence Length & 512 \\
Mixed Precision & bfloat16 \\
\bottomrule
\end{tabular}
\end{table}

evaluation loss of 1.293 (LoRA: $r=16, \alpha =32$) and 1.289, and a evaluation mean token accuracy of about 0.731.
For the first Ministal model ($r=16, \alpha =32$), training was ended after approximately 1.574 epochs and for the second one ($r=32, \alpha =64$) after 2.117 epochs. The first model reaches an evaluation loss of 1.99 with mean token accuracy of 0.594 and the second one 1.978 with a mean token accuracy of 0.614. 
The hyperparameters for the fine-tuning can be seen in \ref{tab:hyperparameters}. The parameters here are chosen following the findings by \cite{huLoraLowrankAdaptation2022a}, which indicate that targeting more weight matrices generally improves performance and thus the LoRA adapters were applied to the query, value, key and output projections of all transformer layers. For the Qwen models, this resulted in roughly 10.09 (0.13\%) respectively 20.19 million (0.27\%) trainable parameters, and for the Ministral models 15.34 (0.19\%) and 30.67 million (0.38\%) trainable parameters.

\textbf{Evaluation Metrics}
To evaluate personality fidelity, accuracy and macro F1 were calculated for each of the four MBTI dimensions, measuring the agreement between the intended personality conditions and the labels inferred by the judging models. Inter-rater reliability was assessed using Krippendorff's $\alpha$, quantifying the agreement between judges. To assess linguistic characteristics, the proportion of English-language output was measured, while lexical diversity was evaluated using Distinct-1 and Distinct-2, representing the proportion of unique unigrams and bigrams, respectively.

\subsection{Qualitative Evaluation and Inter-Rater Agreement}
The LLMs-as-a-Judge approach has produced a large amount of qualitative data and deductive codings, which would be out of scope for this article to analyse in great detail. However, a number of exemplary findings should be demonstrated. Fig. \ref{fig:qual_disagreement} shows how the judging models at times focus on different things about the same text segment, coming to different conclusions: While both Qwen models reasoned that the speaker was reflecting on a hypothetical scenario, which they interpreted as a marker for \textit{Intuition (N)}, the Mistral model instead focused on the aspect that sparked this reflection, which lies in the present and the need to wear a mask, labelling the segment with \textit{Sensing (S)}.

\begin{figure}[htbp]
    \centering
    \begin{tcolorbox}[
    width=0.95\columnwidth,
    colback=gray!5,
    colframe=gray!50,
    boxrule=0.5pt,
    arc=1pt,
    left=5pt,
    right=5pt,
    top=5pt,
    bottom=5pt
]
    \begin{minipage}{0.95\columnwidth}
        \small
        \textbf{Utterance (Person A):} \\
        ``If someone had told me six months ago that we'd have to wear masks at all times, I wouldn't believe them.'' \\[0.7em]
        \textbf{Judge 1 (Qwen3-32B)} $\rightarrow$ \textbf{Intuition (N)} \\
        \textit{Reasoning:} Focuses on future possibilities and abstract scenarios. \\[0.2em]
        \textbf{Judge 2 (Qwen3-30B)} $\rightarrow$ \textbf{Intuition (N)} \\
        \textit{Reasoning:} Reflects on a hypothetical scenario contradicting past expectations. \\[0.2em]
        \textbf{Judge 3 (Mistral-Small-24B)} $\rightarrow$ \textbf{Sensing (S)} \\
        \textit{Reasoning:} Focuses on the present and practical aspects of mask-wearing.
    \end{minipage}
    \end{tcolorbox}
    \caption{Example of inter-rater disagreement caused by differing dimensional interpretations of an identical text segment.}
    \label{fig:qual_disagreement}
\end{figure}

Strong agreement could also be explained by very similar reasoning: For example, the sentence "Hey! Put that thing back up!" is described by every judge as \textit{direct} and \textit{action-oriented}, resulting in a unison \textit{Extraversion (E)} label. At other times models might get conflicting information about the speaking person, for example, when at first strong criticism is uttered ("You really need to stop making excuses for yourself", \textit{Thinking (T)}), but then an affectionate goodbye follows ("Take care, by! Love you lots!", \textit{Feeling (F)}). The latter also giving a hint on a strong limitation of the used dataset: social media posts being written in a certain type of language, which might make the generated utterance fall out of place. A goodbye like that would be much more common on a social media platform than in a work context.

The inter-rater reliability was calculated by applying Krippendorff’s Alpha. Since the Judging Models decided for themselves which text segments to code, they haven't always coded the same ones. For when they have, the agreement looks as shown in table \ref{tab:krippendorff_alpha_1}. Agreement is highest on the J/P dimension, reaching the recommended threshold of 0.800. Agreement on the N/S dimension reaches the 0.667 threshold that allows to draw what Krippendorff calls "tentative conclusions" \cite{krippendorffComputingKrippendorffsAlphareliability2011}, and the other two dimensions miss it slightly. However, the small sample size for any other dimension than I/E means that these results need to be viewed with caution.

Agreement on the final judgement of the speakers is shown in table \ref{tab:master_results}. A few things stand out: first of all, agreement on the text of the baseline models was higher in the F/T and the J/P dimension, and also in the N/S dimension for the Mistral baseline. However, for the I/E dimension and N/S for the Qwen model family, agreement was highest for the model with the smaller LoRA configuration. Second, the $\alpha$ value for the N/S dimension is immensely lower throughout all models than in any other dimension. Lastly, the threshold of $\alpha = 0.8$ was not reached for any model or dimension, making it fairly difficult to draw robust conclusions from these results. 

\begin{table}[h!]
\centering
\caption{Inter-rater reliability (Krippendorff's $\alpha$) evaluated on identical text segments.}
\label{tab:krippendorff_alpha_1}
\begin{tabular}{lcc}
\toprule
\textbf{Dimension} & \textbf{$\alpha$} & \textbf{Identical Segments ($n$)} \\
\midrule
I/E & 0.644 & 57 \\
N/S & 0.693 & 10 \\
F/T & 0.600 & 3  \\
J/P & 0.847 & 9  \\
\bottomrule
\end{tabular}
\end{table}

\subsection{MBTI Fidelity}
The evaluation is conducted on a test set of \(N = 225\) dialogues per generating model. Since each multi-turn dialogue features two interacting personas (Person A and Person B), this corresponds to a maximum of \(N = 450\) individual persona evaluations per model. Due to missing judge outputs for some dialogues, the actual number of evaluated persona instances ranges from 392 to 450 across models.

Table \ref{tab:master_results} shows that across both model families and all four MBTI dimensions, the judges generally achieved the highest classification performance on the outputs generated by baseline models. Applying the interpretation of this measurement as a proxy for the capability of the model to role-play certain personality types, this result shows that fine-tuning does not yield a better performance in doing so compared to the baseline models. 

Depending on the dimension, F1 scores for the baseline models fall between 0.595 (I/E dimension) and 0.820 (F/T) for Ministral and between 0.530 (I/E) and 0.791 (F/T) for Qwen. For both model families, judges were most accurate at classifying the F/T dimension with, followed by the J/P, the N/S and lastly the I/E dimension. This pattern also holds for the fine-tuned models. The F1 scores for the fine-tuned models were generally lower, ranging from 0.446 (LoRA: $r=16$, $\alpha=32$) to 0.738 ($r=16$,$\alpha = 32$) for Ministral and from 0.425 ($r=32$, $\alpha=64$) to 0.777 ($r=16$,$\alpha=32$) for Qwen. 

If higher rank $r$ and increased scaling factor $\alpha$ make it easier for the judging models to detect the displayed personality type is not conclusively inferable from the results. Performance decreased slightly on the I/E and F/T dimensions for both model families, while on N/S it showed small improvements. Results for J/P differed between the two model families, with an improvement for Ministral but a slight decrease for Qwen.

Calculating McNemar's Test to test for significant differences between the model configurations shows that the outputs of baseline models for both model families are significantly more accurately classified in accordance with the supposedly role-played personality, than those generated by the fine-tuned models. Only the F/T dimension in the texts generated by the Qwen model with the smaller LoRA configuration was not significantly more accurately classified by the judges than its baseline.

\begin{table*}[h!]
\centering
\caption{Comprehensive evaluation results: Consensus fidelity (Accuracy, Macro F1), inter-rater reliability (Krippendorff's $\alpha$), and linguistic generation quality (English proportion, Distinct-1/2).}
\label{tab:master_results}
\resizebox{\textwidth}{!}{%
\begin{tabular}{@{} l ccc ccc ccc ccc @{\hskip 1.5em} ccc @{}}
\toprule
\multirow{3}{*}{\textbf{Model Variant}} & \multicolumn{12}{c}{\textbf{Fidelity (Acc, F1) \& Reliability ($\alpha$)}} & \multicolumn{3}{c}{\textbf{Linguistics}} \\
\cmidrule(lr){2-13} \cmidrule(l){14-16}
& \multicolumn{3}{c}{\textbf{I/E}} & \multicolumn{3}{c}{\textbf{N/S}} & \multicolumn{3}{c}{\textbf{F/T}} & \multicolumn{3}{c}{\textbf{J/P}} & \multirow{2}{*}{Eng (\%)} & \multirow{2}{*}{D-1} & \multirow{2}{*}{D-2} \\
\cmidrule(lr){2-4} \cmidrule(lr){5-7} \cmidrule(lr){8-10} \cmidrule(lr){11-13}
& Acc & F1 & $\alpha$ & Acc & F1 & $\alpha$ & Acc & F1 & $\alpha$ & Acc & F1 & $\alpha$ & & & \\
\midrule
\multicolumn{16}{l}{\textbf{Ministral}} \\
Base & \textbf{0.596} & \textbf{0.595} & 0.435 & \textbf{0.620} & \textbf{0.607} & \textbf{0.164} & \textbf{0.820} & \textbf{0.820} & \textbf{0.621} & \textbf{0.760} & \textbf{0.758} & \textbf{0.654} & \textbf{99.97} & 0.0889 & \textbf{0.5450} \\
LoRA ($r=16, \alpha=32$) & 0.462 & 0.462 & \textbf{0.452} & 0.500 & 0.486 & 0.086 & 0.738 & 0.738 & 0.453 & 0.633 & 0.633 & 0.382 & 99.56 & 0.0882 & 0.5160 \\
LoRA ($r=32, \alpha=64$) & 0.449 & 0.446 & 0.425 & 0.524 & 0.517 & 0.080 & 0.729 & 0.729 & 0.487 & 0.687 & 0.687 & 0.483 & 99.43 & \textbf{0.0968} & 0.5383 \\
\midrule
\multicolumn{16}{l}{\textbf{Qwen}} \\
Base & \textbf{0.533} & \textbf{0.530} & 0.428 & \textbf{0.604} & \textbf{0.596} & 0.024 & \textbf{0.791} & \textbf{0.791} & \textbf{0.691} & \textbf{0.729} & \textbf{0.721} & \textbf{0.583} & 88.35 & 0.0812 & 0.4268 \\
LoRA ($r=16, \alpha=32$) & 0.458 & 0.455 & \textbf{0.555} & 0.529 & 0.525 & \textbf{0.045} & 0.778 & 0.777 & 0.540 & 0.673 & 0.672 & 0.457 & \textbf{92.03} & 0.0961 & \textbf{0.5524} \\
LoRA ($r=32, \alpha=64$) & 0.429 & 0.425 & 0.406 & 0.549 & 0.543 & 0.027 & 0.758 & 0.758 & 0.523 & 0.669 & 0.668 & 0.393 & 91.71 & \textbf{0.0965} & 0.5450 \\
\bottomrule
\end{tabular}%
}
\end{table*}

To examine whether personality fidelity varies with the social context, we additionally compare performance across the three scenarios, shown in table. \ref{tab:scenario_f1}. Mean Macro F1 across the four MBTI dimensions shows no consistent monotonic relationship with scenario complexity. While some dimensions vary substantially between scenarios, the direction of these changes differs across models and dimensions. For example, F/T remains comparatively robust across all scenarios, whereas E/I shows consistently lower and more variable performance. Overall, these results suggest that scenario context affects the expression and recognition of individual personality dimensions, but does not systematically determine overall personality fidelity.

\begin{table}[t]
\centering
\caption{Mean Macro F1 across the four MBTI dimensions for each scenario (Low, Medium, and High Urgency).}
\label{tab:scenario_f1}
\begin{tabular}{lccc}
\toprule
\textbf{Model Variant} & \textbf{Low U.} & \textbf{Medium U.} & \textbf{High U.} \\
\midrule
\multicolumn{4}{l}{\textbf{Ministral}} \\
Base & \textbf{0.711} & \textbf{0.690} & \textbf{0.681} \\
LoRA ($r=16, \alpha=32$) & 0.558 & 0.574 & 0.605 \\
LoRA ($r=32, \alpha=64$) & 0.555 & 0.602 & 0.626 \\
\midrule
\multicolumn{4}{l}{\textbf{Qwen}} \\
Base & \textbf{0.678} & \textbf{0.627} & \textbf{0.663} \\
LoRA ($r=16, \alpha=32$) & 0.573 & 0.618 & 0.628 \\
LoRA ($r=32, \alpha=64$) & 0.588 & 0.586 & 0.617 \\
\bottomrule
\end{tabular}
\end{table}

\subsection{Linguistic Characteristics}
To investigate linguistic characteristics of the generated dialogues, the rate of English-language text was calculated using the \textit{langdetect} package for Python \cite{danilakLangdetect2021} and is shown in table \ref{tab:master_results}. The Ministral models consistently showed an English-rate of $>99\%$. However, of the texts generated by the Qwen baseline model only $88.35\%$ were in English or were detected as such; the majority the non-English texts being in Chinese. Fine-tuning raises the English-rate of the Qwen models to $92\%$ for the smaller LoRA configuration respectively $91.7\%$ for the larger one.

Distinct-1, measuring distinct unigrams, and Distinct-2, measuring distinct bigrams, were calculated to compare lexical diversity of the generated text, shown in table \ref{tab:master_results}. For Ministral, while the large LoRA configuration boosts the Distinct-1 value by about 0.008 compared to the baseline and the small configuration, the baseline produces more diverse bigrams compared to the fine-tuned models. For the Qwen models, LoRA increases both values by about 0.015 for Distinct-1 and by about 0.125 for Distinct-2, with the smaller configuration producing the most diverse texts.

\section{Discussion}
The results show that the judging models are better at classifying the texts that were generated by the baseline models than those of the fine-tuned models. Following the logic of this being a proxy measure for the capabilities of the generating models at role-playing different personalities, this implies that the proposed fine-tuning approach does not increase these capabilities compared to the baseline models. This result holds when compared across scenarios. Between the two applied LoRA configurations, the effects were ambiguous and neither could lift the performance above baseline. Judge agreement mostly lingers around the lower $\alpha$ threshold. Only in the N/S dimension agreement in the fidelity analysis showed a strong drop, compared to an unsuspicious $\alpha$ for identical text segments. This points to a general difficulty of the judging models to conclusively classify a speaker with either \textit{Intuition} or \textit{Sensing} characteristic, and possibly to mixed signals in the dialogues.

Using a dataset made up of social media posts for dialogue generation is a major domain shift, even though the posts are of conversational nature. Employing a labelled version of the DailyDialog dataset might not be sufficient to counter this domain shift. The fine-tuning approach of only varying the LoRA parameters \textit{rank} and \textit{$\alpha$} but not also investigating different configurations of target modules, weakens the conclusion that fine-tuning did not improve personality role-playing performance and would have to be the subject of further studies.  

Lastly, LLMs-as-a-Judge evaluation is subject to model-specific biases and stochasticity \cite{guSurveyLLMasaJudge2026}. The low inter-rater agreement observed here highlights that personality-related qualitative judgments are particularly sensitive to evaluator interpretation. Increasing the number and diversity of judges and validating selected codings with human experts are promising directions for future work.

\newpage
\bibliographystyle{IEEEtran}
\bibliography{IEEEabrv,references}

\end{document}